\documentclass[journal]{IEEEtran}

\ifCLASSINFOpdf
\else
   \usepackage[dvips]{graphicx}
\fi
\usepackage{url}

\usepackage{graphicx}

\usepackage{booktabs}

\usepackage{amsmath}

\usepackage{multirow}

\usepackage{cite}

\usepackage{color}

\begin{document}

\title{Cluster-Guided Unsupervised Domain Adaptation for Deep Speaker Embedding}

\author{Haiquan Mao, Feng Hong, \IEEEmembership{Member, IEEE}, and Man-wai Mak, \IEEEmembership{Senior Member, IEEE}
\thanks{This work was in part supported by the National Natural Science Foundation of China under Grant 61971371. {\em (Corresponding author: Man-wai Mak.)} }
\thanks{Haiquan Mao is with Shanghai Acoustics Laboratory, Chinese Academy of Sciences, Shanghai, China, and University of Chinese Academy of Sciences, Beijing, China (e-mail: maohaiquan20@mails.ucas.ac.cn).}
\thanks{Feng Hong is with The Shenzhen Research Institute, The Hong Kong Polytechnic University, Shenzhen, China, and Shanghai Acoustics Laboratory, Chinese Academy of Sciences, Shanghai, China (e-mail: hongfeng@mail.ioa.ac.cn).}
\thanks{Man-wai Mak is with the Department of Electronic and Information Engineering, The Hong Kong Polytechnic University, Hong Kong, China (e-mail: enmwmak@polyu.edu.hk).}}

\markboth{Journal of \LaTeX\ Class Files, Vol. 14, No. 8, August 2015}
{Shell \MakeLowercase{\textit{et al.}}: Bare Demo of IEEEtran.cls for IEEE Journals}
\maketitle

\begin{abstract}
Recent studies have shown that pseudo labels can contribute to unsupervised domain adaptation (UDA) for speaker verification. Inspired by the self-training strategies that use an existing classifier to label the unlabeled data for retraining, we propose a cluster-guided UDA framework that labels the target domain data by clustering and combines the labeled source domain data and pseudo-labeled target domain data to train a speaker embedding network. To improve the cluster quality, we train a speaker embedding network dedicated for clustering by minimizing the contrastive center loss. The goal is to reduce the distance between an embedding and its assigned cluster center while enlarging the distance between the embedding and the other cluster centers. Using VoxCeleb2 as the source domain and CN-Celeb1 as the target domain, we demonstrate that the proposed method can achieve an equal error rate (EER) of 8.10\% on the CN-Celeb1 evaluation set without using any labels from the target domain. This result outperforms the supervised baseline by 39.6\% and is the state-of-the-art UDA performance on this corpus.
\end{abstract}


\IEEEpeerreviewmaketitle

\section{Introduction}
\IEEEPARstart{S}{peaker} verification aims at verifying whether a given speech corresponds to the claimed speaker \cite{ref1}. In recent years, deep neural networks have significantly improved the performance of speaker verification systems \cite{ref2}. However, training these models to achieve good performance requires a large amount of labeled data. As manual labeling is laborious, well-labeled training data in a specific domain are not always available. When a well-trained model is applied to an unseen domain, mismatch between the source domain and the target domain can result in severe performance degradation.

Unsupervised domain adaptation (UDA) addresses the domain-mismatch problem by utilizing labeled source domain data and unlabeled target domain data \cite{ref3, ref4}, leading to performance improvement on the target domain. For better utilization of the unlabeled data and to enhance the feature extraction capability of the resulting network, contrastive self-supervised learning (CSL) has been proposed to increase the similarity of positive training pairs and decrease the similarity of negative pairs in the latent embedding space. Typical examples of CSL are SimCLR \cite{ref5} and MoCo \cite{ref6} in computer vision. These methods also work well in speaker recognition \cite{ref7, ref8, ref9}. Chen {\em et al}. \cite{ref10} proposed a CSL-based domain adaptation approach and achieved promising results on CN-Celeb. However, CSL requires a large batch size to attain its full potential but the batch size is constrained by GPU memory \cite{ref5}. Besides, the performance of CSL is limited by the absence of speaker identity information \cite{ref11}. To avoid these problems, researchers applied clustering-based methods [11]–[13] on unlabeled data to obtain pseudo speaker labels so that the conventional cross-entropy loss can be applied. Because the pseudo labels can be noisy, direct minimization of the cross-entropy loss derived from the pseudo labels could lead to poorly performed speaker embedding networks \cite{ref14}. Therefore, many studies \cite{ref11, ref12, ref13} attempted to purify the pseudo labels. Because cluster quality is highly dependent on the structure of the input data, recent works have focused on training a deep network to map the input data to a latent space where separation of clusters is more manageable \cite{ref15}.

In clustering-based UDA methods, the two main problems are that the number of clusters is hard to determine and the estimated pseudo labels may be incorrect. In this paper, we propose a novel UDA framework that can reduce the impact of these two problems. First, we design the contrastive center loss to fine-tune the network. We use this loss to simultaneously realize intra-class compactness and inter-class separability, which is achieved by reducing the distance between an embedding and its assigned cluster center while enlarging the distance between the embedding and the other cluster centers. The fine-tuning process can not only improve the recognition performance but also make the network extract more cluster-friendly embeddings. Therefore, we can obtain the pseudo-labeled target domain data with less noisy labels. Second, because the supervised training result is related to the proportion of noisy labels, we combine the large labeled source domain data and the pseudo-labeled target domain data, which further reduces the proportion of noisy labels in the training data. Thus, the supervised training on the combined dataset can achieve good performance even though the number of clusters is misestimated. Our primary contributions are as follows: (1) we validate that fine-tuning a network using the contrastive center loss can bring improvements in both signification performance and cluster quality on the target domain data; (2) we propose a cluster-guided UDA framework that can maximally leverage the labeled source domain data and the unlabeled target domain data.

\begin{figure*}
\centering
\centerline{\includegraphics[scale=0.83]{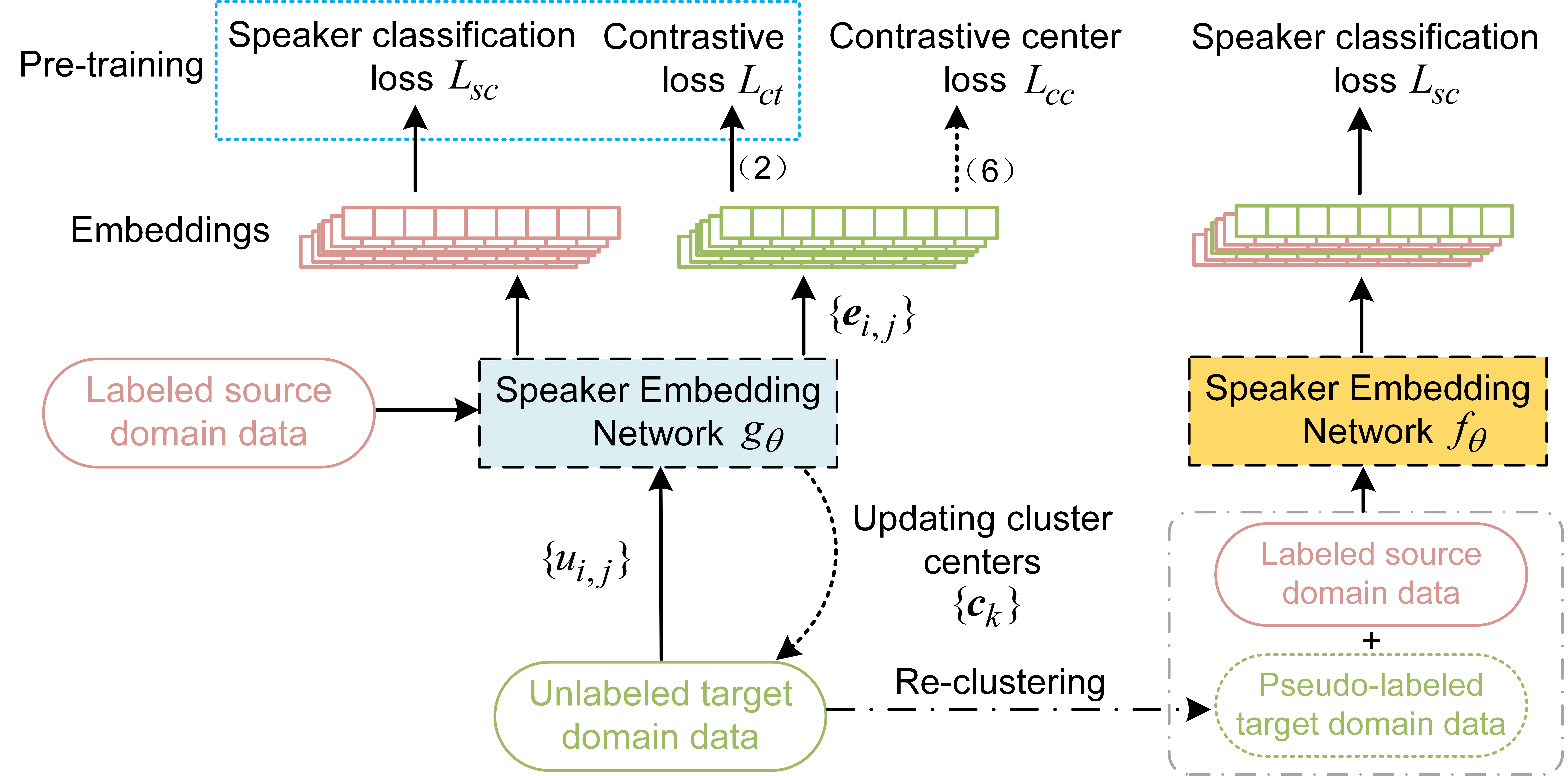}}
\caption{{Overview of the proposed UDA framework. ``Updating cluster centers" means using the speaker embedding network $g_{\theta}$ to extract speaker embeddings of all unlabeled utterances followed by clustering. The obtained cluster assignments and centers are used for computing the contrastive center loss $L_{cc}$. The numbers in parentheses are the equation numbers. $L_{sc}$ denotes speaker classification loss.}}
\end{figure*}

\section{Methods}

\subsection{Training Framework}
As shown in Fig. 1, our UDA framework involves several steps: (1) pre-train a model using the speaker classification loss in the source domain and the contrastive loss in the target domain; (2) extract speaker embeddings and perform clustering on the unlabeled target domain data using the pre-trained model; (3) fine-tune the model with the additional contrastive center loss and update the clusters; (4) use the fine-tuned model $g_{\theta}$ to extract new embeddings, followed by re-clustering to obtain pseudo labels; and (5) conduct supervised training using labeled source domain data and pseudo-labeled target domain data. The well-trained speaker embedding network $f_{\theta}$ is used as the final model.

\subsection{Contrastive Self-supervised Learning}
CSL assumes that two augmented segments from the same utterance form a positive pair, while two segments from different utterances form a negative pair. The main idea is to train the network without using data labels by pulling the samples in positive pairs together and pushing the samples in negative pairs apart in the embedding space.

For each mini-batch $\mathcal{U}$ with $N$ utterances randomly sampled from the unlabeled dataset, it is assumed that these $N$ utterances are from different speakers. For each utterance $u_i$ in the mini-batch, we randomly take two non-overlapping segments of equal length, denoted by $u_{i,1}$ and $u_{i,2}$. Each segment is randomly augmented to emulate various channel characteristics and noise effects, forcing the network to acquire robust speaker representations that are channel-invariant \cite{ref8}. Afterward, all the segments are fed to the speaker embedding network $g_{\theta}$ to extract the speaker embeddings:
\begin{equation}
\boldsymbol{e}_{i, j}=g_{\theta}\left(u_{i, j}\right), \  i=1,2, \ldots, N; \ j=1,2.
\end{equation}

Contrastive loss is designed to maximize the similarity of positive samples and minimize that of negative samples:
\begin{equation}
L_{c t}=-\frac{1}{N} \sum_{i=1}^{N} \log \frac{s\left(\boldsymbol{e}_{i, 1}, \boldsymbol{e}_{i, 2}\right)}{\sum_{m=1}^{N} s\left(\boldsymbol{e}_{i, 1}, \boldsymbol{e}_{m, 2}\right)},
\end{equation}
where $s(\cdot, \cdot)$ denotes a score function that can be Euclidean distance
\begin{equation}
s_1(\boldsymbol{x}, \boldsymbol{y})=\exp \left\{-\frac{1}{\lambda^{2}}\|\boldsymbol{x}-\boldsymbol{y}\|^{2}\right\}
\end{equation}
or cosine similarity
\begin{equation}
s_2(\boldsymbol{x}, \boldsymbol{y})=\exp \{\omega \cos (\boldsymbol{x}, \boldsymbol{y})+b\},
\end{equation}
where $\lambda$, $\omega$, and $b$ are learnable parameters. 

\subsection{Pre-training}
In our strategy, the initial clustering of unlabeled data requires a pre-trained model, which can be obtained either by applying supervised learning on the labeled source domain data or by applying CSL on the unlabeled target domain data. Because Chen {\em et al}. \cite{ref10} have shown that joint training with the labeled source domain data and the unlabeled target domain data can achieve better performance, we performed the same joint training in which the source domain speaker classification loss $L_{sc}$ and the target domain contrastive loss $L_{ct}$ are simultaneously minimized:
\begin{equation}
L_{pre}=L_{sc}+\alpha L_{ct},
\end{equation}
where $\alpha$ is a weighting factor. Here, the speaker classification loss $L_{sc}$ can be the common Softmax loss or some variants of it, such as A-Softmax \cite{ref16}, AM-Softmax \cite{ref17, ref18}, and AAM-Softmax \cite{ref19}.

\subsection{Fine-tuning the Model for Better Clustering}
Generally, the performance of CSL is limited because it does not use speaker labels. An intuitive method is to cluster the unlabeled data to produce pseudo labels and then perform supervised training. However, considering that too much label noise could severely hinder the effectiveness of this strategy, directly using the pseudo-labeled data may not achieve good performance. To reduce the label noise, we dedicatedly train a network for clustering. This network minimizes the joint loss comprising speaker classification loss $L_{sc}$, contrastive loss $L_{ct}$, and contrastive center loss $L_{cc}$ to fine-tune the pre-trained model.

\subsubsection{Contrastive Center Loss}
After extracting the speaker embeddings of all unlabeled utterances, the {\em k}-means algorithm is employed to generate cluster assignments and centers. Generally, discriminative speaker embeddings should have good intra-class compactness and inter-class separability. Therefore, if we force each embedding to be near its assigned cluster center and separate from other cluster centers, {\em k}-means would result in better clustering performance. The contrastive center loss is defined as
\begin{equation}
L_{cc}=-\frac{1}{N} \sum_{i=1}^{N} \log \frac{s\left(\boldsymbol{e}_{i}, \boldsymbol{c}_{y_i}\right)}{\sum_{k=1}^{K} s\left(\boldsymbol{e}_{i}, \boldsymbol{c}_{k}\right)},
\end{equation}
where $\boldsymbol{e}_i$ is the average of $\boldsymbol{e}_{i,1}$ and $\boldsymbol{e}_{i,2}$, $\boldsymbol{c}_{y_i}$ represents the cluster center for $\boldsymbol{e}_i$, and $\boldsymbol{c}_{k}$ represents any cluster center. $K$ denotes the total number of clusters. The contrastive center loss estimates each speaker embedding as the center of the speaker embeddings. Different from other centroid-based losses which require the true speaker labels to obtain the centers, the centers in the contrastive center loss are obtained by clustering. Although there are some incorrect data labels, most cluster centers can be seen as correct for a given embedding. Therefore, the fine-tuning process can also improve the signification performance of the network.

\subsubsection{Joint Loss}
During the fine-tuning process, the speaker classification loss and the contrastive loss constrain the network to maintain the speaker discrimination ability on the target domain, while the contrastive center loss forces the network to map the input data to a more clustering-friendly embedding space. The joint loss is their weighted sum:
\begin{equation}
L_{joint}=L_{sc}+\alpha L_{ct}+\beta L_{cc},
\end{equation}
where $\alpha$ is the same as that in Eq. 5 and $\beta$ is a preset parameter.

\subsubsection{Cluster Updates}
At the beginning, we apply {\em k}-means on the embeddings extracted from the pre-trained model. As fine-tuning continues, the network can extract embeddings more conducive to clustering. Correspondingly, it is necessary to update the cluster assignments and centers. Thus, we perform re-clustering every $P$ epochs during the fine-tuning process. When the contrastive center loss converges, the fine-tuning process ends.

\subsection{Supervised Training on the Combined Dataset}
We use the fine-tuned model $g_{\theta}$ to extract speaker embeddings of the unlabeled target domain data and then produce pseudo labels using the {\em k}-means algorithm. To perform domain adaptation, the labeled source domain data are combined with the pseudo-labeled target domain data, and the total number of classes is their sum. The proportion of noisy labels is low in the combined dataset, which reduces the impact of noisy labels on supervised training. Finally, a new speaker embedding network $f_{\theta}$ is trained with a classification layer and cross-entropy loss on the combined dataset.

\section{Experimental Setup}

\subsection{Datasets}
We conducted experiments on VoxCeleb2 \cite{ref20} and CN-Celeb1 \cite{ref21}, which were used as the source domain data and the target domain data, respectively. VoxCeleb2 is a large-scale English corpus that contains over one million utterances from 5,994 speakers. CN-Celeb1 is a multi-genre Chinese corpus, in which 797 and 200 speakers are reserved for training and evaluation, respectively. Because over 30\% of the utterances in CN-Celeb1 are shorter than 2 seconds, we followed the previous work \cite{ref10} and spliced the short utterances in the training data to utterances longer than 5 seconds. The number of training utterances after splicing is 64,150. The speaker labels of the CN-Celeb1 training data were not used in the UDA framework. Instead, they were only used when verifying the performance of clustering. For data augmentation, the MUSAN corpus \cite{ref22} and room impulse responses (RIR) \cite{ref23} were used for adding additive noise and reverb.

We evaluated the performance of all the training strategies on the CN-Celeb1 evaluation set, using the official trial list that contains 3,484,292 trial pairs in total. In our experiments, we used cosine similarity to score the trial pairs and the performance was evaluated in EER and minimum detection cost function (minDCF).

\subsection{Network Topology and Hyperparameters}
The architecture of the speaker embedding network followed the ECAPA-TDNN in \cite{ref24}. In addition, attentive statistics pooling (ASP) was used as the pooling layer. The channel size was set to 1,024. The embedding dimension of ECAPA-TDNN was set to 192. The supervised training used the additive angular margin loss (AAM-Softmax loss) \cite{ref25} with a margin of 0.2 and softmax prescaling of 30. The network parameters were optimized by an Adam optimizer \cite{ref26}. The initial learning rate was set to 0.001 and decreased by 5\% for each epoch.

All speech files have a sampling rate of 16 kHz. For each file, an 80-dimensional log Mel-spectrogram was extracted from a 25ms window with a 10ms frame shift before being fed to the network. For the supervised training, the duration of each utterance was cropped to 2s. As for the CSL, each utterance in the CN-Celeb1 training set was randomly truncated into two non-overlapping 2-sec segments. Because cosine metric is more relevant and natural with AAM-Softmax and the existing work {\cite{ref27}} has shown that cosine-based similarity metric outperforms Euclidean distance in metric learning objectives, we used cosine similarity to perform k-means and calculate the contrastive loss and the contrastive center loss. The speaker embeddings and centers are length-normalized during the fine-tuning process.

The initial values of $\omega$ and $b$ in Eq. 4 were set to 10 and -5. The weighting coefficients $\alpha$ and $\beta$  in Eq. 7 were set to 1.0. First we set the number of clusters $K$ to 800, and then we varied it to investigate its influence on the performance. We evaluated the performance of the network $g_{\theta}$ every 5 epochs and accordingly $P$ was set to 5. The batch size for supervised training is 256, whereas that for contrastive self-supervised training is 128.

\section{Results and Discussions}

\subsection{Model Performance before Domain Adaptation}\label{formats}
In this section, we report the performance of CSL and compare it with that of supervised training on both the source domain and the target domain. As shown in Table I, although the performance of contrastive self-supervised training is worse than that of supervised training on CN-Celeb1, CSL is still an effective way to learn speaker representations when the data labels are unknown. Besides, due to the domain mismatch and the multi-genre challenge, the results of supervised training on both VoxCeleb2 and CN-Celeb1 are unsatisfactory. The result of supervised training on VoxCeleb2 is considered the baseline for subsequent experiments.

\begin{table}[htbp]
\centering
\caption{Results on CN-Celeb1 without domain adaptation}
\begin{tabular}{llll}
\toprule
Training Data              & Training Mode & EER (\%) & minDCF \\
\midrule
VoxCeleb2                  & Supervised    & 13.40    & 0.56   \\
\midrule
\multirow{2}{*}{CN-Celeb1} & Supervised    & 11.04    & 0.45   \\
                           & CSL           & 13.36    & 0.59   \\
\bottomrule
\end{tabular}
\end{table}

\subsection{The Effectiveness of the Fine-tuned Model for Clustering}
To validate that the fine-tuned model can facilitate clustering, we compared the clustering results using different models for extracting speaker embeddings. To evaluate the cluster quality, we used purity and normalized mutual information (NMI) as the external metrics, which compare the clustering results with the ground truth. The higher purity or NMI indicates better cluster quality. Besides, we used Calinski-Harabasz (CH) indexes \cite{ref28} and Silhouette scores (SS) \cite{ref29} as the internal metrics to evaluate the clustering performance without using any actual labels. A higher CH or SS indicates that the clustering has high intra-class compactness and inter-class separability. The first two lines in Table II reveal that although the supervised training ($L_{sc}$) on the source domain data has poorer performance in terms of EER, it has higher purity and NMI but lower CH and SS. A possible reason is that without speaker labels, the contrastive self-supervised training mistakenly decreases the distance between two utterances belonging to different speakers. Comparing the last two lines, we can see that by adding the contrastive center loss to fine-tune the model, the clustering performance is better in both external and internal metrics.
\begin{table}[htbp]
\centering
\caption{EER and clustering performance of the speaker embedding vectors based on different combinations of training losses}
\begin{tabular}{llllll}
\toprule
Training Loss          & EER (\%)       & Purity         & NMI            & CH               & SS             \\
\midrule
$L_{sc}$               & 13.40          & 0.717          & 0.829          & 91.356           & 0.099          \\
$L_{ct}$               & 13.36          & 0.658          & 0.804          & 130.130 		 & 0.153          \\
$L_{sc}+L_{ct}$        & 10.30          & 0.786          & 0.875          & 100.239          & 0.136          \\
$L_{sc}+L_{ct}+L_{cc}$ & \textbf{9.48}  & \textbf{0.829} & \textbf{0.898} & \textbf{137.544} & \textbf{0.185} \\
\bottomrule

\end{tabular}
\end{table}

Additionally, we further analyze the variations of EER and NMI during the fine-tunning process. As shown in Fig.2, the fine-tuning process can bring improvements in both recognition performance (EER) and clustering performance (NMI), and such improvements are less affected by the misestimation of the total number of clusters.

\begin{figure}
        \center
        \scriptsize
        \begin{tabular}{cc}
                \includegraphics[scale=0.25]{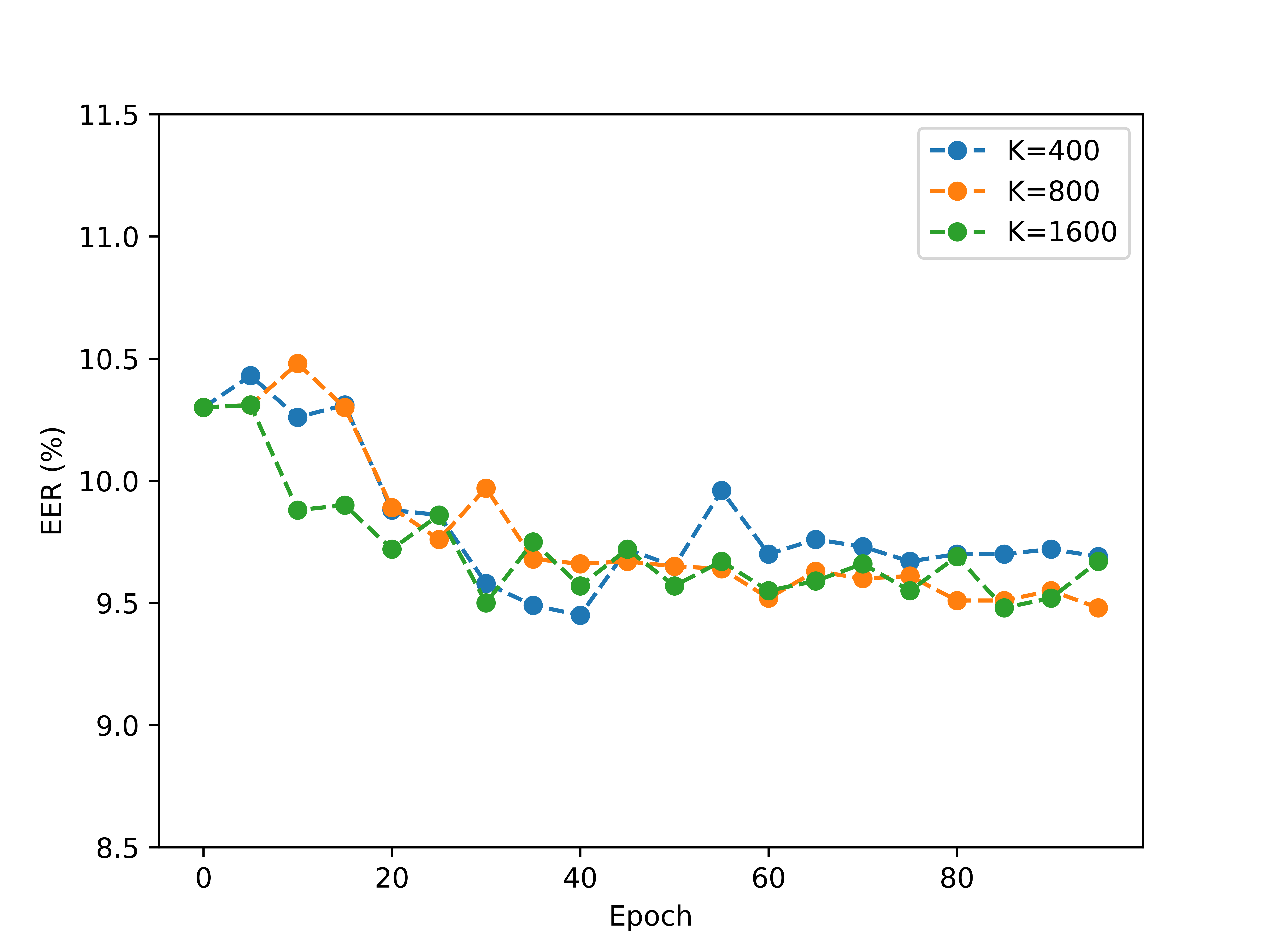}     & \includegraphics[scale=0.25]{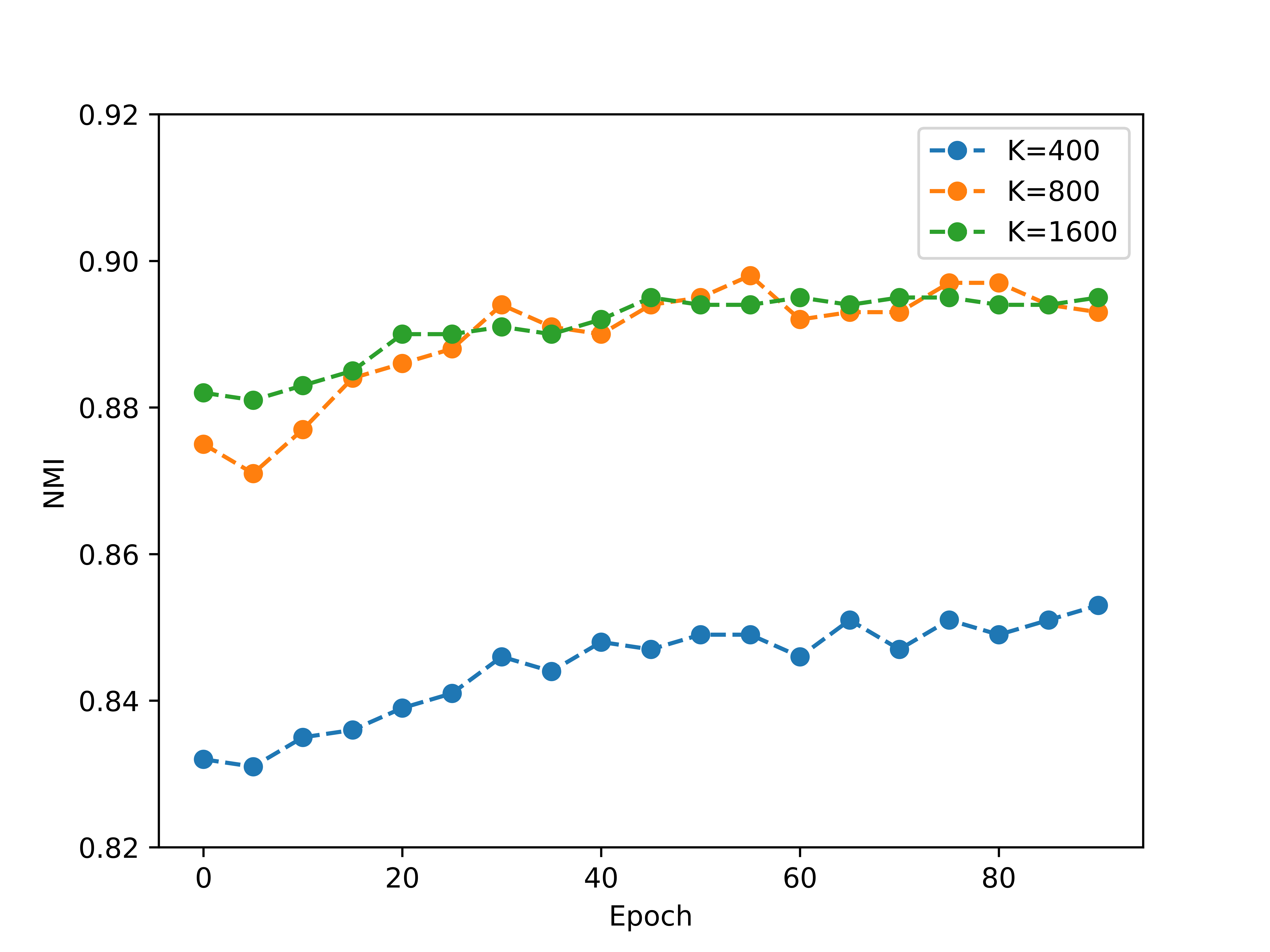}         \\
                (a) EER                              & (b) NMI                                  \\
        \end{tabular}
        \caption{The variations of EER and NMI during the fine-tuning process.}
        \label{figure}
        \vspace{-0.5em}
\end{figure}

\subsection{Performance Comparison with Existing Methods}
Table III compares the performance of our framework with existing methods on the CN-Celeb1 evaluation set. We also varied the number of clusters $K$ to explore its effect on the performance. Evidently, when the true labels are available on both domains, supervised training on the combined dataset outperforms all the existing methods. This result is considered the upper bound performance in our UDA method. When the number of clusters was set to 800, which is close to the actual number of speakers in CN-Celeb1 ($K$=797), our method can achieve 39.6\% performance improvement over the baseline model. This result is very close to the upper bound performance. Although the performance of the models varies with $K$, all of them outperform the baseline model, which indicates that our method brings significant performance improvement even when the number of clusters deviates from the true one.

\begin{table}[htbp]
\centering
\caption{EER Performance of the proposed method and other existing methods on CN-Celeb1}
\begin{tabular}{llll}
\toprule
Method                            & Training Data & Backbone       & EER (\%)      \\
\midrule
Fan {\em et al}. {\cite{ref21}}   & CN1        & i-vector       & 14.24         \\
Li {\em et al}. {\cite{ref30}}    & CN1+CN2    & x-vector       & 12.52         \\
Zhang {\em et al}. {\cite{ref31}} & Vox2       & Rep-A-TMS-TDNN & 10.25         \\
Zeng {\em et al}. {\cite{ref32}}  & CN1+CN2    & ECAPA-TDNN     & 8.93          \\
Chen {\em et al}. {\cite{ref10}}  & Vox2+CN1   & r-vector       & 8.86          \\
Ours                              & Vox2+CN1   & ECAPA-TDNN     & \textbf{8.05} \\
\midrule
Chen {\em et al}. {\cite{ref10}}  & Vox2+CN1.U & r-vector       & 10.20         \\
Ours ($K$=400)                    & Vox2+CN1.U & ECAPA-TDNN     & 8.38          \\
Ours ($K$=800)                    & Vox2+CN1.U & ECAPA-TDNN     & \textbf{8.10} \\
Ours ($K$=1600)                   & Vox2+CN1.U & ECAPA-TDNN     & 8.99          \\
\bottomrule
\multicolumn{4}{p{215pt}}{CN2 means CN-Celeb2 dataset {\cite{ref30}}, which is a supplement of CN-Celeb1 and contains 529485 from 2000 speaker. CN1.U means CN-Celeb1 without data labels.}\\
\end{tabular}
\end{table}

\section{Conclusion}
This paper proposes an effective cluster-guided UDA framework, which consists of fine-tuning the network using contrastive center loss for better clustering and the supervised training on the combined dataset for robust representation learning. The proposed method achieves an EER of 8.10\% on CN-Celeb1 without using any target domain labels, and works well even when the number of clusters deviates from the actual number of speakers.

\end{document}